\documentclass{article}

\PassOptionsToPackage{numbers, compress}{natbib}
 \usepackage[preprint]{nips_2018}

\usepackage[T1]{fontenc}    % use 8-bit T1 fonts
\usepackage{url}            % simple URL typesetting
\usepackage{booktabs}       % professional-quality tables
\usepackage{amsfonts}       % blackboard math symbols
\usepackage{nicefrac}       % compact symbols for 1/2, etc.
\usepackage{microtype}      % microtypography
\usepackage{tabularx}
\usepackage{algorithm}
\usepackage{algorithmic}
\usepackage{epsfig}
\usepackage{amsmath}
\usepackage{multirow}
\usepackage{subfigure}

\title{Multiple Graph Adversarial Learning}

\author{
  Bo Jiang, Ziyan Zhang, Jin Tang and Bin Luo\\
  School of Computer Science and Technology \\
  Anhui University\\
  Hefei, China \\
  \texttt{jiangbo@ahu.edu.cn} \\
}

\begin{document}

\maketitle

\begin{abstract}
Recently, Graph Convolutional Networks (GCNs) have
been widely studied for graph-structured data representation and learning.
However, in many real applications, data are coming with multiple graphs, and it is non-trivial to adapt GCNs to deal with data representation with multiple graph structures.
 One main challenge for multi-graph representation is how to
 exploit both structure information of each individual graph and correlation information across multiple graphs simultaneously.
%The main challenge for multi-graph representation is how to exploit both structure information of intra-graph and correlation information of
% inter-graphs simultaneously.
In this paper, we propose a novel Multiple Graph Adversarial Learning (MGAL) framework for multi-graph representation and learning.
MGAL aims to learn an optimal structure-invariant and consistent representation for multiple graphs in a common subspace via a novel adversarial learning framework, which thus incorporates both structure information of intra-graph and correlation information of
 inter-graphs simultaneously.
Based on MGAL, we then provide a unified network for semi-supervised learning task.
 % by employing an adversarial learning way.
Promising experimental results demonstrate the effectiveness of MGAL model.
\end{abstract}

\section{Introduction}

Graph representation and learning which aims to represent each graph node as
a low-dimensional feature vector is a fundamental problem in pattern recognition and machine learning area.
Recently, Graph Convolutional Networks (GCNs) have been widely studied for graph representation and learning~\cite{duvenaud2015convolutional,atwood2016diffusion,adaptive_GCN,velickovic2017graph}.
These methods can be categorized into spatial and spectral methods.
Spatial methods generally define graph convolution operation  by designing an operator on node neighbors while
spectral methods usually define graph convolution operation based on spectral analysis of graphs.
For example,
Monti et al. \cite{monti2017geometric} present mixture model CNNs (MoNet) for graphs. % and provide a unified generalization of CNN architectures on graphs.
Velickovic et al. \cite{velickovic2017graph} present Graph Attention Networks (GAT) for graph semi-supervised learning. % that aggregates the features of neighboring nodes.
% Some recent works have also been proposed~\cite{velickovic2017graph}.
%they generally  define graph convolution operation based on spectral representation of graphs.
% For example,
For spectral methods,
Bruna et al. \cite{bruna2014spectral} propose to define graph convolution  based on eigen-decomposition of graph Laplacian matrix.
Henaff et al. \cite{henaff2015deep} further introduce a spatially constrained spectral filters. % with smooth coefficients in order to make them spatially localized.
Defferrard et al. \cite{defferrard2016convolutional} propose to approximate the spectral filters based on Chebyshev expansion. % of graph Laplacian to avoid the high computational complexity of eigen-decomposition.
Kipf et al. \cite{kipf2016semi}  propose a more simple Graph Convolutional Network (GCN) based on first-order approximation of spectral filters.
%Velickovic et al. \cite{velickovic2017graph} present Graph Attention Networks (GAT) by designing a novel attention layer that aggregates the features
%of the neighboring nodes for semi-supervised learning. %weighted by some learnable importance.
%Li et al. \cite{velickovic2017graph} propose optimal GCNs, in which the graph is learned optimally by traditional distance metric learning.

However, in many applications, data are coming with multiple graphs, which is known as multi-graph learning~\cite{nie2016parameter}.
In this paper, we focus on multiple graphs
that contain the common node set but with multiple different edge structures.
%It is usually assumed that each
%individual graph captures the partial information.
The aim of our multi-graph representation is to find a consistent node representation across all of graphs.
One main challenge for this problem is how to
effectively integrate multiple graph (edge) structures together in graph node representation.
%, i.e., how to
%correlation information of
% inter-graphs simultaneously. while  structure information of each individual graph in node representation.
The above existing GCNs generally cannot be used to deal with multiple graphs.
%exploit
%
%i) structure information of each individual intra-graph and ii) correlation information of
% inter-graphs simultaneously.
To generalize GCNs to multiple graphs, one popular way is to use some heuristic fusion strategy and transform multi-graph learning to traditional single graph learning~\cite{pham2017column,simonovsky2017dynamic,schlichtkrull2018modeling}.
%For example,
%Pham et al, propose to first train different convolution parameters and obtain representations  for different graphs and then obtain the representation in each layer by averaging the representations of them.
%Similar idea has also been proposed in work~\cite{888}.
%
%** proposes a Relational GCN (R-GCN) which takes a similar idea in knowledge graphs by further training different weights for different graphs.
However, since the fusion process is usually independent of graph representation/learning, this strategy may lead to weak optimal solution.
In addition, some works propose to conduct the convolution operation on multiple graphs
by sharing the common convolution parameters across different graphs~\cite{duvenaud2015convolutional,atwood2016diffusion,zhuang2018dual}.
 This mechanism can propagate some information/knowledge across multiple graphs.
 However,
 the learned representations of individual graphs in these methods are still not guaranteed to be consistent explicitly.
 In this paper, we propose  a novel Multiple Graph Adversarial Learning (MGAL) framework for multi-graph representation and learning.
MGAL aims to learn an optimal structure-invariant and thus consistent representation for multiple graphs in a common subspace via an adversarial learning architecture. %, and thus incorporates both structure information of intra-graph and correlation information of  inter-graphs simultaneously.
% To the best of our knowledge,  it is the first effort to develop adversarial learning framework for multi-graph learning.
Based on MGAL, we further provide a unified network for multi-graph based semi-supervised learning task.
Overall, the main contributions of this paper are summarized as follows:
\begin{itemize}
  \item  We propose a novel Multiple Graph Adversarial Learning (MGAL) framework for multi-graph representation and learning.
  The proposed MGAL is a general framework which allows to generalize any learnable/parameterized graph representation models to deal with multiple graphs.
  \item  We present a unified  network for semi-supervised learning task based on multi-graph representation.
  \item We develop a general generative adversarial learning architecture (`multiple generators + one discriminator') to address the general multi-view representation and learning problem.
\end{itemize}
Experimental results on several datasets demonstrate the effectiveness and benefits of the proposed MGAL model and semi-supervised learning method. % significantly outperforms the state-of-the-art graph neural network models on semi-supervised learning tasks.

\section{Problem Formulation}

\textbf{Graph representation.}
Let $G(X,A)$ be an attributed graph where $X=(x_1,x_2\cdots x_n)\in \mathbb{R}^{n\times d}$ denotes the collection of node features
and %$n$ data feature vectors in $d$ dimensional space.
$A \in \mathbb{R}^{n\times n}$  encodes the pairwise relationships (such as similarities) between node pairs. The aim of graph representation is to
learn a latent representation $Z = f(X,A;\Theta)$ where $Z=(z_1\cdots z_n)\in \mathbb{R}^{n\times k}, k\leq d$  in a low-dimensional space that
takes in both graph structure $A$ and node content $X$ together.
One kind of popular way is to use Graph Convolutional Networks (GCNs)~\cite{kipf2016semi,velickovic2017graph,henaff2015deep,monti2017geometric} which provide a unified framework to define the representation function $f(X,A;\Theta)$.
Based on representation $Z$, we can then conduct some learning tasks, such as node classification, clustering and semi-supervised classification, etc.
% represent both graph structure $A$ and node content $X$ simultaneously.

\textbf{Multi-graph representation.} In many real applications,
data are coming with multiple graphs, which is known as multi-graph representation/learning problem.
In this paper, we focus on multiple graphs
that contain the common node set and the same node content but with multiple different edge structures.
Formally, given $G(X,\mathcal{A})$ with $X\in \mathbb{R}^{n\times d}$ denoting node content and $\mathcal{A}=\{A^{(1)},A^{(1)}\cdots A^{(m)}\}$ representing the multiple edge structures, the aim of our multi-graph representation
is to learn a consistent latent representation $Z = h(X,\mathcal{A};\Theta)$, where $Z\in \mathbb{R}^{n\times k}, k\leq d$ in a low-dimensional space that
takes in multiple graph structures $\mathcal{A}$ and node content $X$ together.
Based on representation $Z$, we can then conduct some learning tasks, such as node classification, clustering and semi-supervised learning, etc.
In this paper, we focus on semi-supervised learning.
 The main challenge for multi-graph representation is how to
 exploit the information of each individual graph $A^{(v)}$ while take in the correlation cue among multiple graphs simultaneously in final representation $Z$.
 A simple and direct way to use multiple graphs is to average them to a new one and then put it into the standard GCNs model.
Obviously, since the graph average process is independent of graph learning, this may lead to weak local optimal solution.

\section{Related Works}

\textbf{Multiple graph convolutional representation.}
To generalize GCNs to multiple graphs, one can first obtain the representation $Z^{(v)} = f(X,A^{(v)};\Theta^{(v)})$ for each graph $A^{(v)}$ individually by using GCNs and then concatenate or average them together to obtain the final representation $Z$.
Obviously, this two-stage strategy  neglects the correlation information  among different graphs.
To overcome this issue, some works propose to conduct ensemble of different representations in middle layers of GCNs to
share/communicate some information across different graphs in learning process~\cite{pham2017column,simonovsky2017dynamic,schlichtkrull2018modeling}.
 Since the fusion in each layer is independent of graph learning, this strategy may still lead to weak optimal representation.

%Pham et al.~\cite{pham2017column} propose to first train different convolution parameters to obtain representations  for different graphs and then obtain the representation as the output in each layer by averaging them.
%Similar idea has also been proposed in works~\cite{simonovsky2017dynamic,schlichtkrull2018modeling}.
%
%Schlichtkrull et al.~\cite{schlichtkrull2018modeling} propose a Relational GCN (R-GCN) which takes a similar idea in knowledge graphs by further training different weights for different graphs.

Another kind of works propose to conduct the convolution operation on multiple graphs
by sharing the common convolution parameters across different graphs, i.e.,
$Z^{(v)}=f(X,A^{(v)};\Theta)$~\cite{duvenaud2015convolutional,atwood2016diffusion,zhuang2018dual}.
This mechanism can propagate some correlation information across graphs $A^{(v)}, v=1\cdots m$ via the common parameters $\Theta$.
%Duvenaud et al. \cite{duvenaud2015convolutional}
%propose a convolutional neural networks (Neural FPs) that can operate directly on multiple graphs
%by sharing the parameters across different graphs.
%Atwood and Towsley \cite{atwood2016diffusion} propose Diffusion-Convolutional Neural Networks (DCNNs) in which the parameters can be shared across graphs of arbitrary sizes.
%Zhuang and Ma \cite{zhuang2018dual} propose a dual graph convolutional network by using
%the common parameters on dual graphs.
However, although the parameters $\Theta$ are common for different graphs, the learned representations $\{Z^{(1)}\cdots Z^{(m)}\}$ are still not guaranteed to be well consistent because each $Z^{(v)}=f(X,A^{(v)};\Theta)$ is determined by not only parameters $\Theta$ but also individual graph $A^{(v)}$.

\textbf{Adversarial learning model.}
 Our multiple graph adversarial learning model is inspired by
 Generative Adversarial Network (GAN)~\cite{goodfellow2014generative}, which consists of a generator $\mathcal{G}$ and
 a discriminator $\mathcal{D}$.
 The generator is trained to generate
 the samples to convince the discriminator while the discriminator aims to
 discriminate the samples returned by generator.
 Recently, adversarial learning has been explored in graph representation tasks.
 Wang et al.~\cite{wang2018graphgan} propose a graph representation model with GANs (GraphGAN).
 Dai et al.~\cite{dai2018adversarial} propose an adversarial network embedding (ANE), which employs
 the adversarial learning to regularize graph representation.
Pan et al.~\cite{pan2018adversarially} also propose an adversarially regularized graph autoencoder model for graph embedding.

Different from previous works, our aim in this paper is to derive a general adversarial learning framework  for multiple graph representation and learning.
 To the best of our knowledge,  it is the first effort to develop adversarial learning framework for multi-graph learning problem.
\begin{figure*}[!htbp]
\centering
  \includegraphics[width=0.995\textwidth]{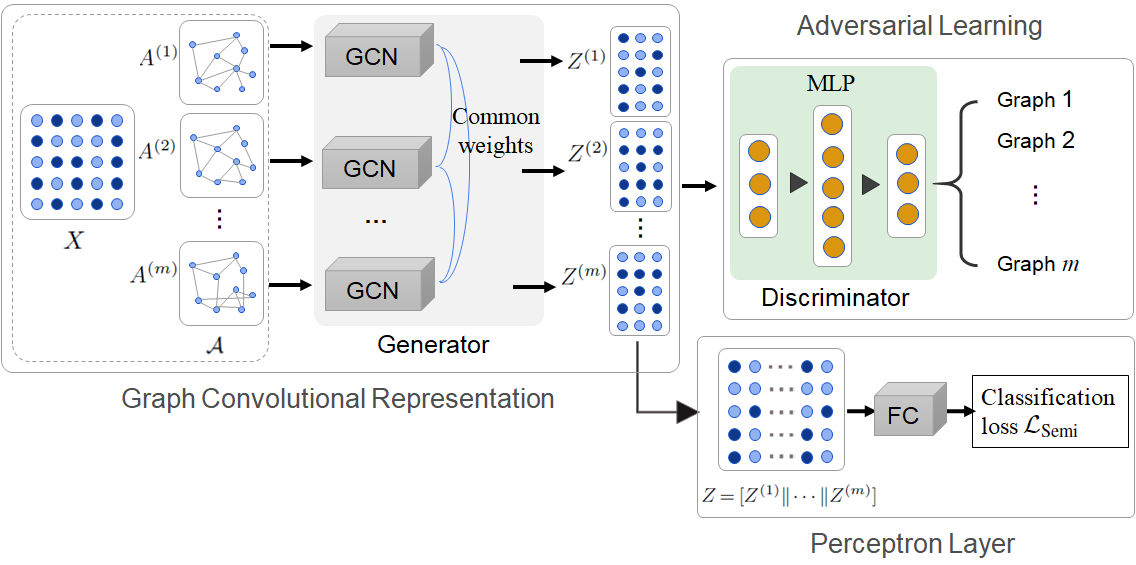}
\caption{Architecture of the proposed multiple graph adversarial learning. It consists of
three main modules, i.e., graph convolutional representation, adversarial learning network and  perceptron layer for node label prediction.}
\label{fig:framework}       % Give a unique label
\end{figure*}

\section{The Proposed Model}

\subsection{Overall Framework}

Given an input feature matrix $X$ and associated multiple graph structures $\mathcal{A}=\{A^{(1)}\cdots A^{(m)}\}$,
our aim is to learn a consistent latent representation $Z = h(X,\mathcal{A};\Theta)$
and then conduct graph node semi-supervised classification in a unified network model.
Figure 1 demonstrates the overall architecture of our Multiple Graph Adversarial Learning (MGAL) which consists of
three main modules, i.e., graph convolutional representation, adversarial learning module and  perceptron layer for node label prediction.
\begin{itemize}
  \item \textbf{Graph convolutional representation.} We employ several
  graph convolutional layers to learn a latent low-dimensional representation $Z^{(v)}$ for each individual graph by incorporating both structure of graph $A^{(v)}$ and node features $X$ together.
  \item \textbf{Adversarial learning.} The adversarial learning module aims to enforce the latent representations $Z^{(v)}$ of graphs to be consistent
      and generate a kind of structure-invariant representations for different graphs ${A}^{(v)}$ in a common subspace.
%The discriminator is a graph classifier that tries to discriminate
%whether the current latent representation $Z^{(v)}$ comes from the $v$-th graph or other graphs. % convolutional layer.
  \item \textbf{Perceptron layer.} We obtain the final representation $Z$ by concatenating $Z^{(v)}$ together and  use a  perceptron layer to predict the
  label for each node, and thus conduct semi-supervised classification for graph nodes.
\end{itemize}
%

%vIn the following, we introduce them respectively in detail.

\subsection{Graph Convolutional Representation}

The graph convolutional representation aims to learn a latent representation $Z^{(v)}=f(X,A^{(v)};\Theta)$ for nodes of each graph by exploring both structure of graph $A^{(v)}$ and node feature $X$ together.
%
%In many real applications, data come with multiple graphs. %, such as co-saliency detection task discussed latter.
%Therefore, it is necessary to extend the above GCN model to deal with multiple graphs.
%Formally, let $\mathcal{G}(\mathcal{X},\mathcal{A})=\{G(X^{(1)}, A^{(1)})\cdots G(X^{(N)}, A^{(N)})\}$ be
%a set of $N$ graphs with arbitrary sizes, where $X^{(v)}=(x^{(v)}_1\cdots x^{(v)}_{n_v})\in \mathbb{R}^{n_{v}\times d_{v}}$ and $A^{(v)} \in \mathbb{R}^{n_v\times n_v}$
%denote the node feature collection and  adjacency matrix of the $v$-th graph, respectively.
%To deal with multiple graph data $\mathcal{G}(\mathcal{X},\mathcal{A})$,
In this paper, we propose to employ graph convolutional networks (GCNs)~\cite{kipf2016semi,duvenaud2015convolutional,atwood2016diffusion} which have been widely studied in recent years. The main property of GCNs architecture is that it can represent both graph structure $A^{(v)}$ and node content $X$ in a unified framework.
% GCNs extends the operation of convolution to graph data.
Generally, given a graph $G(X,A)$, GCNs conduct layer-wise propagation in hidden layers as
\begin{align}\label{EQ:general_gcn}
H^{(v)}_{l+1} = f_l(H^{(v)}_{l},A^{(v)};\Theta_{l})
\end{align}
where $l=0,1\cdots L-1, v=1,2\cdots m$ and $L$ is the number of layers. $H^{(v)}_{0}=X\in \mathbb{R}^{n\times d}$ is the initial input feature matrix.
 $H^{(v)}_{l}$ denotes the input feature map of the $l$-th layer  and  $H^{(v)}_{l+1}$ is the
output feature representation after conducting graph convolutional operation on graph $A^{(v)}$. Parameters $\Theta =\{\Theta_0\cdots \Theta_{L-1}\}$ are layer-specific trainable weight matrices needing to be learned.
We can thus use the final output of GCNs as a compact representation  for each individual graph $A^{(v)}$ as
\begin{equation}\label{aa}
Z^{(v)}=f(X,A^{(v)};\Theta)=H^{(v)}_{L}
\end{equation}
Many methods have been proposed to provide various kinds of graph convolution operator $f_l$ in recent years.
In this paper, we employ the widely used spectral convolution function~\cite{kipf2016semi} defined as
%
%\begin{small}
\begin{align}\label{EQ:general_gcn}
f_l(H^{(v)}_{l},A^{(v)};\Theta_{l}) = \sigma\big({D^{(v)}}^{-\frac{1}{2}}\tilde{A}^{(v)}{D^{(v)}}^{-\frac{1}{2}}H^{(v)}_{l}\Theta_{l}\big)\nonumber
\end{align}
%\end{small}
%
where
 $\tilde{A}^{(v)} = A^{(v)} +I$ and $D^{(v)} = \sum^n_{j=1} \tilde{A}^{(v)}_{ij}$. $I$ is the identity
 matrix and $\sigma$ denotes an activation function, such as $\textrm{ReLU}(\cdot) = \max(0,\cdot)$.

%%
%\begin{align}\label{EQ:layer_gcn}
%X^{(v)}_{l+1} = \sigma({D^{(v)}}^{-\frac{1}{2}}\tilde{A}^{(v)}{D^{(v)}}^{-\frac{1}{2}}X^{(v)}_{l}W_{l})
%\end{align}
%%
%where $ l=0,1\cdots L-1$.
%Here $\tilde{A}^{(v)} = D^{-1/2}A^{(v)}D^{-1/2}+I$ and $D = \sum^n_{j=1} A^{(v)}_{ij}$.
%The matrices  $\mathcal{W}=\{W_1,W_2\cdots W_{K-1}\}, W_k\in \mathbb{R}^{d_k\times d_{k+1}}$ are layer-specific trainable weight matrices to conduct feature extraction and representation.
%Function $\sigma$ denotes an activation function, such as $\textrm{ReLU}(\cdot) = \max(0; \cdot)$.

\noindent \textbf{Remark.} Note that, in our graph convolutional architecture, the weight parameters $\Theta=\{\Theta_0\cdots \Theta_L\}$ of hidden layers
for different graphs  are shared which can thus propagate some information/knowledge across different graphs for node representation, as suggested in works~\cite{duvenaud2015convolutional,atwood2016diffusion}.

%$A^{(1)} A^{(2)}  A^{(m)} Z^{(1)} Z^{(2)}  Z^{(m)} $
%Obviously, this separate learning way fails to  model the interactions  or correlations  across multiple
%feature representations $\mathcal{X}$ and graphs $\mathcal{A}$ when they are in the same domain, such as
%multiple related images for co-saliency detection (as discussed in \S 4 in detail).
%For multiple graph learning, it is desirable to propagate the information/knowledge from one graph to another graph and seek one consistent and boosting learning
%across all graphs.
%In this paper, we achieve this purpose by sharing the common weight matrices of hidden layers across different graphs and  conduct the layer-wise propagation in hidden layers as
%%
%\begin{align}\label{EQ:layer_gcn}
%X^{(v)}_{k+1} = \mathrm{ReLU}(\tilde{A}^{(v)}X^{(v)}_{k}W_{k}), \  \ k=0,1\cdots K-1
%\end{align}
%%
%where $\mathcal{W}=\{W_1,W_2\cdots W_{K-1}\}$ are the common weight matrices of different hidden layers across multiple graphs.

\subsection{Adversarial Learning}

% This strategy can propagate information/knowledge across multiple graphs via the common parameters $\Theta$.
% and thus obtains a consistent representation and learning by integrating the information of multiple graphs simultaneously.
As mentioned in \S 2, for our multi-graph learning problem, it is necessarily to generate a consistent representation across different graphs in a common subspace.
However, the above learned representations $Z^{(v)}$ (Eq.(2)) can not be guaranteed to be consistent because $Z^{(v)}$ is determined by not only the common shared parameters $\Theta$  but also individual graph $A^{(v)}$.
To overcome this issue, %In order to further enforce the latent representations $Z^{(v)} = f(X,A^{(v)};\Theta)$ of different graphs to be consistent in a common subspace,
we design an adversarial learning module, which consists of a generator and a discriminator~\cite{goodfellow2014generative}.
% This is achieved by an adversarial training module.
 Our adversarial model is built on a standard multi-layer perceptron (MLP), which
 acts as a discriminator to distinguish whether the representation $Z^{(v)}$ is generated from
 graph $A^{(v)}$ or some other graph $A^{(u)}$ where $u\neq v$.

\textbf{Generator} $\mathcal{G}(X,A^{(v)};\Theta)$  aims to generate a kind of
structure-invariant/consistent representations $Z^{(v)}$ for graphs $A^{(v)}$ in
a common subspace. In this paper, we employ the above GCNs as our generator module, i.e.,
\begin{equation}\label{**}
Z^{(v)}=\mathcal{G}(X,A^{(v)};\Theta) = f(X,A^{(v)};\Theta)
\end{equation}
For each node representation, we denote
\begin{equation}\label{**}
z^{(v)}= \mathcal{G}(x,A^{(v)};\Theta) = f(x,A^{(v)};\Theta)
\end{equation}
where $x\in X$ is the input feature vector of a graph node and $z^{(v)}\in Z^{(v)}$ is the corresponding generated representation.
The generator is supervised and optimized by both cross-graph discrimination loss and  node
(semi-supervised) classification loss. %, as shown latter.
 The cross-graph discrimination loss minimizes the gap among different graph representations (as shown in Eq.(5)) while
 the classification loss can separate the representations of graph nodes for node semi-supervised classification task (as shown in Eq.(7)). %\S ** in detail.

%In this paper, we focus on ** with the same node content.
%
%When the node feature are different
%
%across multiple
%feature representations $\mathcal{X}$ and graphs $\mathcal{A}$ when they are in the same domain,
%

%which is defined as
%%
%\begin{equation}\label{EQ:triplet_loss}
%\mathcal{L}_{Reg} = \sum_v\sum_u \|Z^{(v)} - Z^{(u)}\|_F
%\end{equation}
%%
%where $Z^{(v)}, Z^{(u)}$ denote the representations generated from graph $A^{(v)}$ and $A^{(u)}$, respectively.

% consistent or indistinctive representation $Z^{(v)}$ under different graphs $A^{(v)}$.

\textbf{Discriminator} $\mathcal{D}(Z)$ aims to discriminate the representation $Z$ obtained from the generator. It is built on a standard multi-layer perceptron (MLP) and
defined as a multi-class classifier. It outputs a class label indication vector $\mathcal{D}(Z)=(\mathcal{D}^{(1)}(Z)\cdots\mathcal{D}^{(m)}(Z))$ in which $\mathcal{D}^{(v)}(Z)$ represents the probability of representation $Z$ generating from the $v$-th graph $A^{(v)}$ and $\sum^m_{v=1} \mathcal{D}^{(v)}(Z) = 1, \mathcal{D}^{(v)}(Z)\geq 0$.

In our adversarial learning,
generator $\mathcal{G}$ and discriminator $\mathcal{D}$
act as two opponents.
Generator $\mathcal{G}$ would try to
generate an indistinctive representation $Z^{(v)}$ for each graph $A^{(v)}$,
while discriminator $\mathcal{D}$, on the contrary,
 would try to discriminate whether the representation $Z^{(v)}$ is generated from the $v$-th graph or some other graph.
This can be achieved by optimizing the following cross-entropy loss $\mathcal{L}_{\mathrm{MGAL}}$ as
\begin{align}\label{ss}
&\min_{\mathcal{G}}\max_{\mathcal{D}} \mathcal{L}_{\mathrm{MGAL}}=\frac{1}{m}\sum^m_{v=1} \Big(\mathbb{E}_{x}\log\big[\mathcal{D}^{(v)}(\mathcal{G}(x,A^{(v)};\Theta))\big]\nonumber\\
&+\textstyle  \frac{1}{m-1}\sum^m_{u,u\neq v} \mathbb{E}_{x}\log\big[1-\mathcal{D}^{(u)}(\mathcal{G}(x,A^{(v)};\Theta))\big]\Big)
\end{align}

\textbf{Remark.}
In our MGAL, since the convolutional parameters $\Theta$  are shared for multiple graphs $G(X, A^{(v)})$,
thus only one generator is designed.
Note that, here one can also use multiple generators $\mathcal{G}(X, A^{(v)}, \Theta^{(v)})$ with each generator conducting on each individual graph respectively.
This `multiple generators + one discriminator' adversarial learning provides a new feasible architecture to address the general multi-graph (view) representation and learning tasks.

\subsection{Perceptron Layer}

After adversarial training, all the representation $Z^{(v)}, v=1,2\cdots m$ are lied in a common low-dimensional space.
In the final perceptron layer, we can use one of them or aggregate them for the final node representation.
In this paper, %we use the  ensemble of $\{Z^{(1)}\cdots Z^{(m)}\}$ in our final representation.
% we aim to combine the latent consistent representations $Z^{(v)}$ of graphs together to conduct our final graph node (semi-supervised) classification.
% In particular,
we first aggregate $\{Z^{(1)}\cdots Z^{(m)}\}$ by concatenating them  as $Z=[Z^{(1)}\|Z^{(2)}\cdots \|Z^{(m)}]\in \mathbb{R}^{n\times md}$ where $\|$ denotes the concatenation operation
% both $\{X_K^{(1)}\cdots X_K^{(N)}\}$ and $Y_{K}$ together
 and
then employ a fully-connected layer to conduct label prediction as
\begin{equation}\label{EQ:final_gcn}
U = \mathrm{softmax} (ZW)
\end{equation}
where $W\in \mathbb{R}^{md\times c}$ is a trainable weight matrix and $c$ denote the number of node classes.
The final output $U\in \mathbb{R}^{n\times c}$ denotes the
predicted label vectors for graph nodes.
Note that, here one can also use a graph convolutional layer to conduct label prediction as
\begin{equation}\label{EQ:final_gcn}
U = \mathrm{softmax} (\bar{A}ZW)
\end{equation}
where $\bar{A}=\frac{1}{m}\sum^m_{v=1}A^{(v)}$.
For semi-supervised learning,
we aim to minimize the following cross-entropy loss $\mathcal{L}_{\textrm{Semi}}$ over all the labeled nodes $L$, i.e.,
 \begin{equation}
\mathcal{L}_{\textrm{Semi}} = -\sum\nolimits_{i\in L} \sum^c\nolimits_{j=1} {Y}_{ij}\log{U}_{ij}
 \end{equation}
where ${L}$ indicates the set of labeled nodes and ${Y}_{i\cdot}, i\in L$ denotes the corresponding label indication vector for the $i$-th labeled node, i.e.,
\begin{equation}
{Y}_{ij}=
\begin{cases}
1& \text{if node $i$ belongs to $j$-th class}\\
0& \text{otherwise}
\end{cases}
\end{equation}

Figure 2 demonstrates the 2D t-SNE \cite{Geoffrey2017Visualizing} visualization of the representation output by GCN on each individual graph $A^{(v)}$ (denote as GCN(v)) and MGAL on multiple graph $\mathcal{A}$ respectively on {MSRC-v1} dataset~\cite{winn2005locus} (as shown in Experiments).
Different colors denote different classes. Intuitively, one can observe that the data of different classes are distributed more clearly  in MGAL representation, which demonstrates the benefits of MGAL  on conducting multi-graph representation and learning.

%\begin{algorithm}[tb]
%   \caption{MGAL Algorithm}
%   \label{alg:example}
%\begin{algorithmic}
%   \STATE {\bfseries Input:} data $x_i$, size $m$
%   \REPEAT
%   \STATE Initialize $noChange = true$.
%   \FOR{$i=1$ {\bfseries to} $m-1$}
%   \IF{$x_i > x_{i+1}$}
%   \STATE Swap $x_i$ and $x_{i+1}$
%   \STATE $noChange = false$
%   \ENDIF
%   \ENDFOR
%   \UNTIL{$noChange$ is $true$}
%\end{algorithmic}
%\end{algorithm}
%
%%
\begin{figure*}[!htbp]
\centering
  \includegraphics[width=0.98\textwidth]{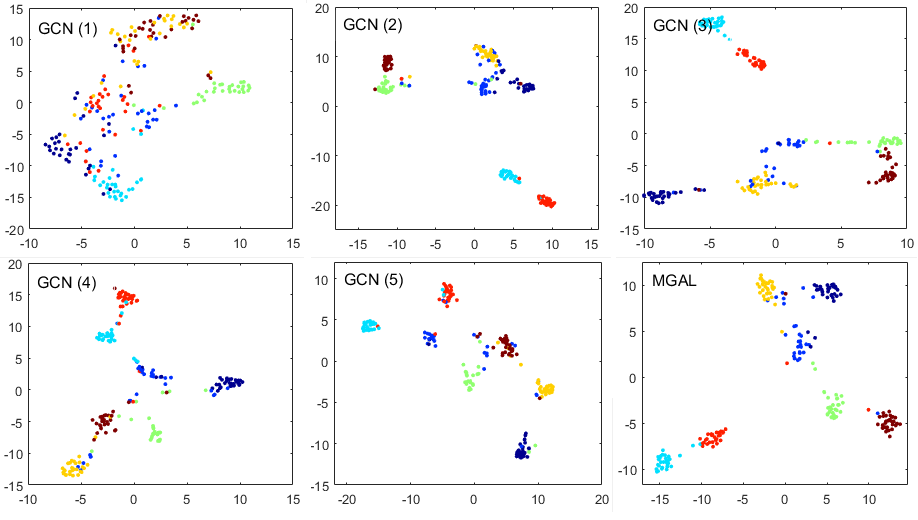}
\caption{2D t-SNE \cite{Geoffrey2017Visualizing} visualizations of the representation based on GCN and MGAL, respectively on MSRC-v1 dataset. Different classes are marked by different colors. Note that, comparing with traditional single graph GCN, the data distribution is demonstrated more clearly in our multi-graph representation. }
\label{fig:framework}       % Give a unique label
\end{figure*}
\begin{table*}[!htp]\small
\centering
\caption{Comparison results of semi-supervised classification on  Handwritten numerals, Caltech101-7, MSRC-v1 and CiteSeer datasets, respectively. The best performance are marked by bold. }
\centering
\begin{tabular}{c||c|c|c||c|c|c}
  \hline
  \hline
  Dataset& \multicolumn{ 3}{c||}{Handwritten numerals} & \multicolumn{ 3}{c}{Caltech101-7}\\
  \hline
   Ratio of label & 10\% & 20\% & 30\% & 10\% & 20\% & 30\% \\
  \hline
  GCN(1) &85.88$\pm$0.91 &87.51$\pm$0.77 & 87.65$\pm$0.60 & 83.55$\pm$0.81 & 84.12$\pm$1.26 & 84.18$\pm$1.37 \\
  GCN(2) & 93.32$\pm$1.41 & 95.33$\pm$0.49 & 95.66$\pm$0.51 & 82.09$\pm$0.40 & 83.59$\pm$0.59 & 85.45$\pm$1.39 \\
  GCN(3) & 95.14$\pm$0.66 & 96.71$\pm$0.47 &97.17$\pm$0.51 &84.92$\pm$0.78 &86.18$\pm$2.34 &86.56$\pm$1.85 \\
  GCN(4) & 95.26$\pm$0.93 & 97.19$\pm$0.22 & 97.46$\pm$0.50 & 93.20$\pm$1.01 & 95.06$\pm$0.59 & 95.76$\pm$0.77  \\
  GCN(5) &84.61$\pm$0.38 &96.71$\pm$0.48 &86.40$\pm$0.64 &91.64$\pm$1.91 &92.47$\pm$3.04 &94.18$\pm$2.24 \\
  GCN(6) & 71.01$\pm$1.87 & 71.44$\pm$1.69 & 73.03$\pm$1.28 & 90.45$\pm$1.98 & 92.36$\pm$2.09 & 94.27$\pm$0.87  \\
  GCN-M & 95.89$\pm$0.67 & 95.85$\pm$2.48 & 97.38$\pm$0.63 & 90.01$\pm$2.44 & 92.42$\pm$2.87 & 93.87$\pm$1.55\\
  Multi-GCN & 96.89$\pm$0.35 & 97.47$\pm$0.38 & 97.89$\pm$0.29 & 93.75$\pm$2.32 & 95.60$\pm$0.82 & 95.28$\pm$3.01\\
  AMGL & 91.82$\pm$0.55 & 94.15$\pm$0.32 & 95.65$\pm$0.73 & 88.83$\pm$0.69 & 92.79$\pm$0.34 & 94.62$\pm$0.42\\
  MLAN & 97.32$\pm$0.23 & 97.48$\pm$0.19 & 97.91$\pm$0.28 & 93.52$\pm$0.58 & 94.84$\pm$0.19 & 95.46$\pm$0.18\\
  \hline
  MGAL & \textbf{97.64$\pm$0.39} & \textbf{98.25$\pm$0.34} & \textbf{98.48$\pm$0.42} & \textbf{95.00$\pm$1.07} & \textbf{96.41$\pm$0.81} & \textbf{97.14$\pm$0.44}  \\
  \hline
  \hline
  Dataset & \multicolumn{ 3}{c||}{MSRC-v1} & \multicolumn{ 3}{c}{CiteSeer}\\
  \hline
  Ratio of label  & 10\% & 20\% & 30\% & 10\% & 20\% & 30\% \\
  \hline
  GCN(1) &42.75$\pm$11.2 &45.09$\pm$4.45 & 45.00$\pm$7.38 & 71.08$\pm$1.83 & 75.62$\pm$1.14 & 77.95$\pm$1.59 \\
  GCN(2) & 77.69$\pm$8.15 & 83.98$\pm$2.47 & 83.43$\pm$5.45 & 73.38$\pm$2.89 & 77.44$\pm$1.16 & 79.15$\pm$0.96 \\
  GCN(3) & 81.21$\pm$4.52 & 84.84$\pm$1.65 &89.57$\pm$2.15 & - & - & - \\
  GCN(4) & 70.66$\pm$2.77 & 75.65$\pm$2.87 & 76.00$\pm$4.49 & - & - & -  \\
  GCN(5) & 71.10$\pm$4.64 &78.88$\pm$5.39 &79.57$\pm$2.84 & - & - & - \\
  GCN-M & 80.77$\pm$4.13 & 86.21$\pm$3.37 & 90.29$\pm$2.15 & 72.84$\pm$2.32 & 77.93$\pm$1.49 & \textbf{80.13$\pm$1.82}\\
  Multi-GCN & 83.30$\pm$2.21 & 88.70$\pm$1.69 & 89.29$\pm$2.43 & 72.04$\pm$1.37 & 76.15$\pm$1.42 & 77.69$\pm$1.73\\
  AMGL & 82.29$\pm$2.43 & 89.35$\pm$2.09 & 90.21$\pm$1.08 & 64.78$\pm$1.58 & 68.00$\pm$1.41 & 70.92$\pm$2.93\\
  MLAN & 83.66$\pm$2.38 & 87.92$\pm$1.18 & 89.47$\pm$1.68 & 63.24$\pm$6.54 & 59.49$\pm$6.72 & 60.47$\pm$6.16\\
  \hline
  MAGL & \textbf{88.68$\pm$0.66} & \textbf{89.93$\pm$2.40} & \textbf{91.29$\pm$1.65} & \textbf{75.18$\pm$1.64} & \textbf{78.84$\pm$1.57} & 78.77$\pm$1.25  \\
  \hline
  \hline
\end{tabular}
\end{table*}
\section{Experiments}

To evaluate the effectiveness of the proposed MGAL and semi-supervised learning method, we implement it and compare it with some other methods on four datasets.

\subsection{Datasets}

We test our MGAL on four datasets including MSRC-v1~\cite{winn2005locus}, Caltech101-7~\cite{li2015large,nie2016parameter}, Handwritten numerals~\cite{AsuncionNewman2007} and CiteSeer~\cite{AsuncionNewman2007}.
The details of these datasets and their usages in our experiments are introduced below.

\textbf{MSRC-v1} dataset~\cite{winn2005locus} contains 8 classes of 240 images.
Following~\cite{nie2016parameter}, in our experiments, we select 7 classes including tree,
building, airplane, cow, face, car and bicycle.
Each class contains 30 images.
Following the experimental setting in work~\cite{nie2016parameter},
five graphs are obtained for this dataset by using five different kinds of visual descriptors, i.e., 24 Color Moment, 576 Histogram of Oriented Gradient, 512 GIST, 256 Local Binary Pattern and 254 Centrist features.

\textbf{Caltech101-7} dataset~\cite{li2015large,nie2016parameter} is an object recognition data set containing
101 categories of images. We follow the experimental setup of previous work \cite{nie2017multi} and
select the widely used 7 classes (Dolla-Bill, Face, Garfield, Motorbikes, Snoopy, Stop-Sign and Windsor-Chair) and obtain 1474 images in our experiments.
Six neighbor graphs are obtained for this dataset by using six different kinds of visual feature descriptors including 48 dimension Gabor feature, 40 dimension wavelet moments (WM), 254 dimension CENTRIST feature, 1984 dimension HOG feature, 512 dimension GIST feature, and 928 dimension LBP feature.

\textbf{Handwritten numerals}~\cite{AsuncionNewman2007} dataset contains 2,000 data points for 0 to 9 ten digit
classes and each class has 200 data points.
We construct six graphs by using six published feature descriptors including 76 Fourier coefficients of
the character shapes, 216 profile correlations, 64 Karhunen-love coefficients, 240 pixel averages in
$2\times 3$ windows, 47 Zernike moment and morphological features.

\textbf{CiteSeer} dataset~\cite{AsuncionNewman2007} consists of 3,312 documents  on scientific publications, which can be further classified into six classes, i.e.,
Agents, AI, DB, IR, ML and HCI.
For our multi-graph learning,
two graphs are built up by using a 3,703-dimensional vector representing
whether the key words are included for the text view and
the other 3279-dimensional vector recording the citing relations
between every two documents, respectively~\cite{nie2016parameter}.
\begin{figure*}[htbp]
\centering
  \includegraphics[width=0.99\textwidth]{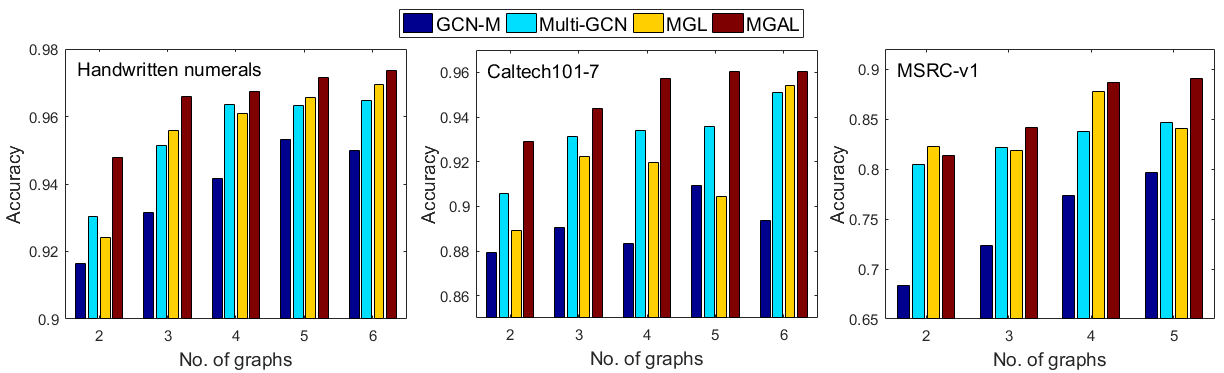}
\caption{Comparison results on different numbers of graphs. }
\label{fig:framework}       % Give a unique label
\end{figure*}

\subsection{Experimental Setup}

\textbf{Parameter setting.}
% Similar to \cite{kipf2016semi},
For generator, we use a three-layer graph convolutional network and the number of units in each hidden layer is set to 64 and 16, respectively.
For discriminator, we use a four-layer full connection network with the number of
units in each layer is set to 16, 64, 16 and $m$ respectively where $m$ denotes the number of graphs, as discussed in \S 4.3.
%
%We present additional experiments on performance of GDENs with different hidden unit numbers and convolutional layers in \S 4.4.
%
We train our generator for a maximum of 500 epochs by using Adam \cite{Adam} with learning rate 0.005.
We train our discriminator for a maximum of 500 epochs by using Stochastic Gradient Descent (SGD)  with learning rate 0.01.
All the network weights are initialized using Glorot initialization~\cite{glorot2010understanding}. We stop training if the validation loss does not decrease for 50 consecutive epochs.

%For Cora, Citeseer and Pubmed datasets,
%we follow the experimental setup in work ~\cite{kipf2016semi,velickovic2017graph}.

\textbf{Data setting.} For all datasets, we select 10\%, 20\% and 30\% nodes as label data per class and
use the remaining data as unlabeled samples.
For unlabeled samples, we also use 5\% nodes for validation purpose to determine the convergence criterion, as suggested in~\cite{kipf2016semi}, and use the remaining 85\%, 75\% and 65\% nodes respectively as test samples.
All the accuracy results are averaged over 5 runs with different data splits.

\textbf{Baselines.}  We compare our method (MGAL) against state-of-the art methods as follows,

\textbf{GCN(v)} that conducts traditional graph convolutional network~\cite{kipf2016semi} on the $v$-th graph $A^{(v)}$ and node content features $X$.\\
\textbf{GCN-M} that conducts traditional graph convolutional network~\cite{kipf2016semi} on the averaged graph $\bar{A}=\frac{1}{m}\sum^m_{v=1}A^{(v)}$ and node features $X$.\\
%\textbf{R-GCN}~\cite{schlichtkrull2018modeling} that trains different parameters for different graphs and then  aggregate them together in the middle layers of network.\\
\textbf{Multi-GCN} that first learns representation for multiple graphs $A^{(v)}$ by using/sharing the common parameters, as suggested in work~\cite{duvenaud2015convolutional}, and then select the representation $Z^{(v)}$ with the lowest training loss function for the final multi-graph representation. \\
\textbf{MGL} that removes the adversarial learning module in our MGAL network. We implement it as a baseline to demonstrate the effectiveness of the proposed adversarial learning.

In addition, we also compare our method with some other recent traditional multiple graph learning and semi-supervised learning methods including:

\textbf{AMGL}~\cite{nie2016parameter} is a parameter-free auto-weighted multiple graph learning and label propagation for semi-supervised learning problem. \\
\textbf{MLAN}~\cite{nie2017multi} is a multi-view learning model for semi-supervised classification. % that conducts traditional GCN on the averaged graph
%
% to determine the convergence criterion, as suggested in~\cite{kipf2016semi}.
%%Note that, for graph based semi-supervised setting, we use all of the nodes in our training.
%
%We implement our GDEN with three versions, i.e.,
%%label smoothness regularization.
%1) GDEN-RWR that utilizes random walk with restart (RWR) based diffusion operator in GDEN;
%2) GDEN-Lap that utilizes graph Laplacian diffusion operator in GDEN;
%3) GDEN-NLap that utilizes normalized Laplacian diffusion operator in GDEN.
%The optimal parameter $\lambda$ in GDEN-RWR, GDEN-Lap and $\gamma$ in GDEN-NLap are determined based on validation loss values.

%
\begin{figure}[!htb]
\centering
  \includegraphics[width=0.7\textwidth]{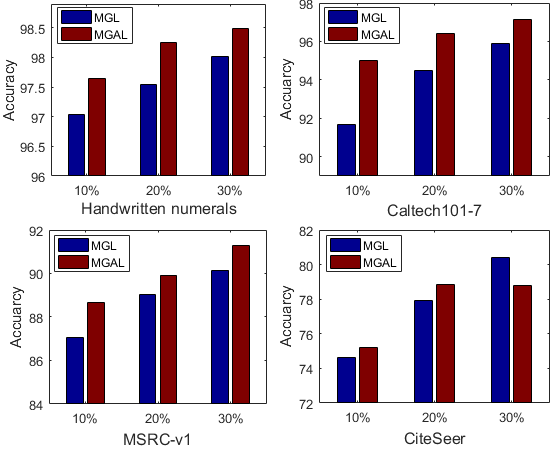}
\caption{Comparison results of our MGAL with the baseline MGL. Note that, MGAL performs better than MGL, indicating the effectiveness of the proposed adversarial learning. }
\label{fig:framework}       % Give a unique label
\end{figure}

\subsection{Comparison Results}

Table 1 summarizes the comparison results of semi-supervised classification on four datasets.
% Table 2 summarizes the comparison results on four widely used image datasets. The best results are marked as bold in Table 1 and 2.
%
Here, one can note that
(1) The proposed MGAL performs obviously better than traditional single GCN method conduced on each individual graph $A^{(v)}$. This clearly demonstrates the benefit of the
proposed multi-graph representation and semi-supervised learning on multiple graphs.
(2) Comparing with the other baseline method GCN-M and Multi-GCN,
the proposed MGAL returns the best performance, which indicates the effectiveness of the proposed multi-graph representation architecture.
(3) MGAL performs better than recent multiple graph learning and semi-supervised learning method {AMGL}~\cite{nie2016parameter}
and {MLAN}~\cite{nie2017multi}, which indicates the benefit of the proposed multi-graph learning and semi-supervised learning model. %GLCN on graph data representation and learning.

Figure 3 shows the classification accuracy of the proposed MGAL method across different numbers of graphs.
For each graph number, we conduct semi-supervised experiments on all possible graph groups and then
compute the average performance. Here, one can note that,
(1) as the number of graphs increases, MGAL obtains better learning performance.
It clearly demonstrates the desired ability of MGAL  on multiple graph integration for robust
learning which is the main issue for multi-graph learning task. (2) Our MGAL performs consistently better than other baseline methods on different sizes of graph set, which
further indicates the advantage of the proposed MGAL.

Figure 4 shows the classification accuracy of the proposed MGAL method
comparing with the baseline method MGL on four datasets.
One can note that, MGAL generally performs better than MGL on all experiments, which clearly indicates the desired benefits of the proposed method by incorporating the adversarial learning architecture in  multi-graph representation and learning.

%\subsection{Algorithms}
%
%If you are using \LaTeX, please use the ``algorithm'' and ``algorithmic''
%environments to format pseudocode. These require
%the corresponding stylefiles, algorithm.sty and
%algorithmic.sty, which are supplied with this package.
%Algorithm~\ref{alg:example} shows an example.

% \section*{Acknowledgements}

\section{Conclusion}

 This paper proposes a novel Multiple Graph Adversarial Learning (MGAL) framework for multi-graph representation and learning.
MGAL aims to learn an optimal structure-invariant and thus consistent representation for multiple graphs in a common subspace, and thus incorporates both structure information of intra-graph and correlation information of
 inter-graphs simultaneously.
Based on MGAL, we then provide a unified network for semi-supervised learning task.
Promising experimental results demonstrate the effectiveness of the proposed MGAL model.
The proposed MGAL is a general framework which allows to generalize any learnable/parameterized graph representation models to deal with multiple graphs.
  % \item We develop a general generative adversarial learning architecture (`multiple generators + one discriminator') to address the general multi-view representation and learning problem.
\bibliographystyle{ieee}
\bibliography{nmfgm_MGAL}

\end{document}